\documentclass{article}
 
\PassOptionsToPackage{numbers, sort&compress}{natbib}
\usepackage[preprint]{neurips_2026}

\usepackage[utf8]{inputenc} 
\usepackage[T1]{fontenc}    
\usepackage{hyperref}       
\usepackage{url}            
\usepackage{booktabs}       
\usepackage{amsfonts}       
\usepackage{nicefrac}       
\usepackage{microtype}      
\usepackage{xcolor}         
\usepackage[table]{xcolor}
\usepackage{graphicx}
\usepackage{amsmath} 
\usepackage{threeparttable}
\usepackage{algorithm}
\usepackage{algpseudocode}
\usepackage{makecell}
\usepackage{enumitem}
\usepackage{multirow}
\usepackage{array}
\usepackage{pifont}
\usepackage{wrapfig}
\usepackage{xspace}

\newcommand{\cmark}{\ding{51}}
\newcommand{\xmark}{\ding{55}}
\newcommand{\pname}{RFWalk\xspace}

\title{Self-Supervised Doppler-Guided RF Odometry}

\author{%
  Kunzhe Song \quad Jialuo Du \quad Jingkai Lin \quad Huacheng Zeng \\
  Michigan State University \\
  \texttt{\{songkunz,dujialuo,linjingk,hzeng\}@msu.edu} \\
}

\begin{document}

\maketitle

\begin{abstract}
Radio-frequency (RF) sensing enables robot ego-motion estimation in darkness, smoke, and fog, yet RF odometry remains challenging because multipath and moving reflectors make cross-frame matching highly ambiguous. Existing methods often rely on auxiliary inertial sensors or ground-truth pose supervision. We present RFWalk, a self-supervised, RF-only odometry framework that models dense correspondences across range-azimuth frames as random walks on a space-time graph and learns them from unlabeled sequences through temporal cycle consistency. To rule out correspondences that are cycle-consistent but physically implausible, we introduce a Doppler consistency loss that constrains correspondence-induced motion using radial velocity measured by the RF sensor itself. Ego-motion is then recovered through robust $\mathrm{SE}(2)$ registration that suppresses uncertain correspondences and those arising from independently moving objects. Without pose labels or auxiliary sensors, RFWalk remains competitive with supervised baselines in static scenes and reduces absolute trajectory error by 58\% in dynamic scenes with moving pedestrians.
\end{abstract}

\section{Introduction}

Reliable ego-motion estimation is a fundamental capability for autonomous robots, underpinning localization, mapping, and navigation.
Cameras and LiDAR have become the dominant sensing modalities for odometry~\citep{campos2021orb, engel2017direct, zhang2014loam}, yet their performance can degrade substantially in darkness, smoke, fog, and dust.
Radio-frequency (RF) sensing, particularly millimeter-wave (mmWave) radar, provides a complementary sensing modality that is robust to illumination changes and airborne obscurants while remaining compact and inexpensive enough for mobile robots~\citep{guan2020through, lu2020see, adib2015capturing, zhao2018through}.
These properties have motivated growing interest in RF-based odometry and simultaneous localization and mapping (SLAM)~\citep{lu2020milliego, sie2024radarize, lai2025rf}.

Despite these advantages, estimating ego-motion from RF observations alone remains challenging.
The core difficulty lies in establishing reliable correspondences between consecutive RF frames.
Compared with optical images or LiDAR point clouds, range-azimuth (RA) heatmaps have coarse angular resolution and are heavily affected by multipath propagation and specular reflections, producing unstable or spurious structures that may vanish or shift across frames.
In feature-poor or repetitive environments, multiple candidate matches can be geometrically indistinguishable.
Moving reflectors such as pedestrians further complicate the problem, since correspondences on independently moving objects do not follow the rigid ego-motion shared by the static scene.

Existing RF odometry methods follow several strategies to address these challenges, but each leaves part of the problem unresolved.
Scan-matching approaches~\citep{adolfsson2021cfear, prabhakara2023high} align consecutive RF observations geometrically, yet remain vulnerable to multipath, coarse angular resolution, and geometric degeneracy.
Radar-inertial systems~\citep{kramer2020radar, doer2020ekf, lu2020milliego, zhuang20234d} improve motion observability by incorporating an IMU, at the cost of additional sensing, calibration, and synchronization.
Doppler-based approaches~\citep{kellner2013instantaneous, sie2024radarize} exploit radar-measured radial velocity as a direct motion cue, but typically assume that most reflectors are static.
Moreover, learning-based components in these pipelines~\citep{prabhakara2023high, sie2024radarize} require ground-truth trajectories or co-located LiDAR supervision, both of which are costly to collect at scale.
As a result, accurate RF-only odometry without pose supervision remains an open problem, particularly in dynamic environments.

\begin{figure}
    \centering
    \includegraphics[width=\linewidth]{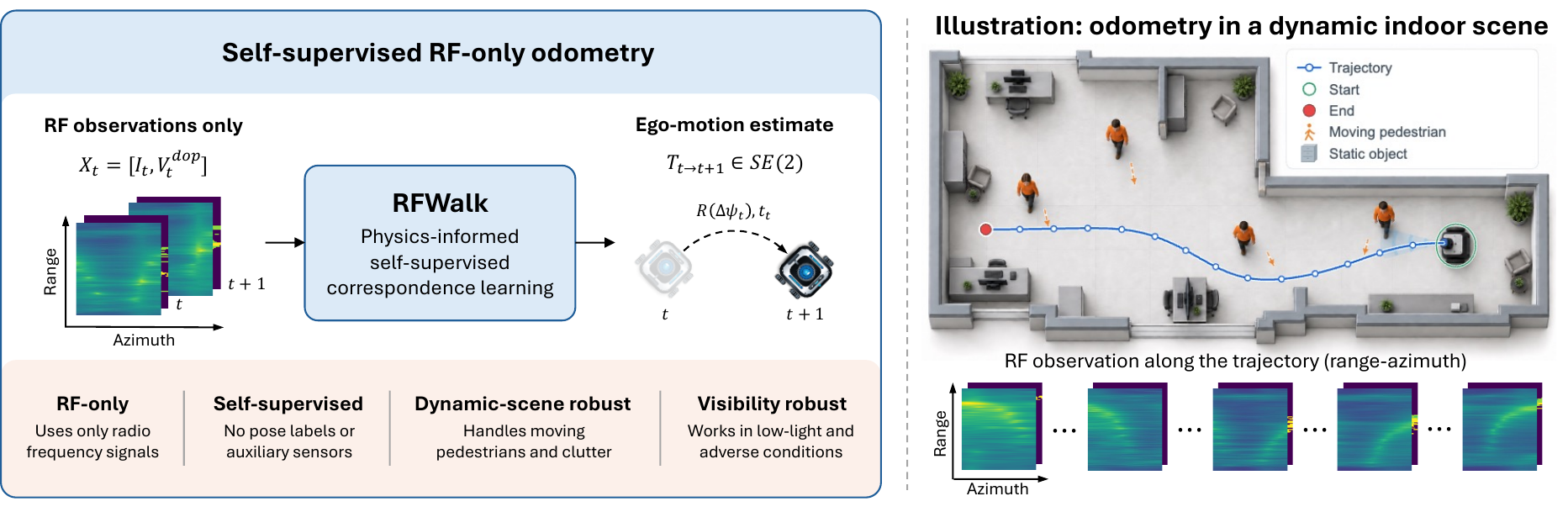}\vspace{-0.10in}
    \caption{Self-supervised RF-only odometry.
    Left: from consecutive RF frames alone, \pname learns cross-frame correspondences without pose labels or auxiliary sensors and recovers the ego-motion $T_{t\rightarrow t+1}$.
    Right: illustration of RF-only odometry in a dynamic indoor scene with moving pedestrians.}
    \label{fig:intro}
\end{figure}

In this work, we introduce \pname, a self-supervised odometry framework that relies solely on RF observations and formulates ego-motion estimation as physics-informed cross-frame correspondence learning (Fig.~\ref{fig:intro}).
Rather than regressing poses directly, \pname models dense correspondences between RA frames as random walks on a space-time graph~\citep{jabri2020space, shrivastava2024self}, learns these correspondences from unlabeled RF sequences, and recovers ego-motion from the resulting correspondence field.
Neither ground-truth poses nor auxiliary sensors are required during training or inference.

The key idea is to combine two complementary sources of self-supervision intrinsic to the RF signal.
Temporal cycle consistency encourages correspondences that persist across time, while Doppler measurements provide an independent physical constraint on the radial displacement implied by each correspondence.
Unlike prior Doppler-based methods, we do not use Doppler to estimate ego-velocity directly; instead, we use it to determine whether each local cross-frame association is physically plausible.
This constraint applies to both static and moving reflectors and penalizes incorrect matches that remain cycle-consistent in repetitive or geometrically ambiguous scenes.
We further enforce consistency between composed multi-step correspondences and their direct counterparts to reduce drift, and recover motion through robust $\mathrm{SE}(2)$ registration that suppresses uncertain or independently moving correspondences.
Together, these components turn ambiguous RF matching into physically grounded, self-supervised motion estimation.

We evaluate \pname on the public Radarize benchmark~\citep{sie2024radarize} and on the \pname dataset, which we collect using a 60\,GHz mmWave ISAC platform in indoor scenes with moving pedestrians.
Without pose supervision or auxiliary sensing, \pname remains competitive with supervised baselines in static scenes and reduces absolute trajectory error by 58\% relative to the strongest baseline in dynamic scenes.
Its advantage becomes more pronounced as scene dynamics increase.
From the least to the most dynamic scenes, its error grows by less than 1\,m, compared with 2.3--8.2\,m for the supervised baselines.
These results suggest that physics-informed correspondence learning offers a promising path toward scalable RF odometry from unlabeled sequences.

We summarize our main contributions as follows.
\begin{itemize}[leftmargin=*,nosep]
    \item We formulate RF odometry as physics-informed, self-supervised correspondence learning, enabling dense cross-frame associations to be learned from unlabeled RF sequences without ground-truth poses or auxiliary sensors.

    \item We introduce Doppler consistency supervision, which converts per-cell radial velocity into a local physical constraint on correspondence-induced displacement. This constraint applies to both static and moving reflectors and suppresses physically incompatible matches that cycle consistency alone cannot rule out.

    \item We enforce compositional consistency across multi-step correspondences and recover motion through confidence-aware robust $\mathrm{SE}(2)$ registration, improving long-horizon stability and robustness to dynamic clutter.
\end{itemize}

\section{Method}

\subsection{RF Observation Representation}

We consider RF odometry in an RF-only setting, where only RF measurements are available during both training and inference. 
At each time step $t$, we represent the RF observation as a two-channel RA tensor
\begin{equation}
    \mathbf{X}_t = \left[ \mathbf{I}_t, \mathbf{V}_t^{\mathrm{dop}} \right] \in \mathbb{R}^{N_r \times N_\theta \times 2},
\end{equation}
where $\mathbf{I}_t$ is the RA log-power map and $\mathbf{V}_t^{\mathrm{dop}}$ is a signed radial-velocity map containing one scalar Doppler velocity for each RA cell. Here, $N_r$ and $N_\theta$ denotes the numbers of range and azimuth bins, respectively. Let $N = N_r N_\theta$ be the number of RA cells. We index each RA cell by $i \in \{1,\dots,N\}$ and associate it with the Cartesian coordinate of its bin center,
$\mathbf{p}_i = [r_i \cos \theta_i \,, r_i \sin \theta_i]^\top \in \mathbb{R}^2$.
Thus, each cell jointly encodes the reflected RF strength at a physical spatial location and the instantaneous radial motion observed along its line of sight.

We use a scalar radial-velocity map instead of retaining all Doppler FFT bins for two reasons. First, radial velocity is an explicit physical quantity that can be directly related to correspondence-induced radial displacement, enabling the physics-informed supervision introduced in Sec.~\ref{method:doppler} without requiring the network to infer the relevant Doppler component from a spectrum. Second, the number, spacing, and velocity range of Doppler bins vary substantially across radar waveforms and configurations. Collapsing the spectrum into metric radial velocity therefore provides a consistent two-channel input interface after platform-specific signal processing, while avoiding sensor-specific changes to the network input dimensionality. Given two consecutive frames $\mathbf{X}_t$ and $\mathbf{X}_{t+1}$, our goal is to estimate the relative planar motion
\begin{equation}
    \mathbf{T}_{t \rightarrow t+1} = \bigl(\mathbf{R}(\Delta\psi_t), \mathbf{t}_t\bigr) \in \mathrm{SE}(2),
\end{equation}
where $\mathbf{R}(\Delta\psi_t) \in \mathbb{R}^{2\times2}$ is the relative rotation and $\mathbf{t}_t \in \mathbb{R}^2$ is the translation. 
The inter-frame interval $\Delta t$ is obtained from RF timestamps.

\subsection{Cross-Frame Correspondence}

\begin{figure}
    \centering
    \includegraphics[width=\linewidth]{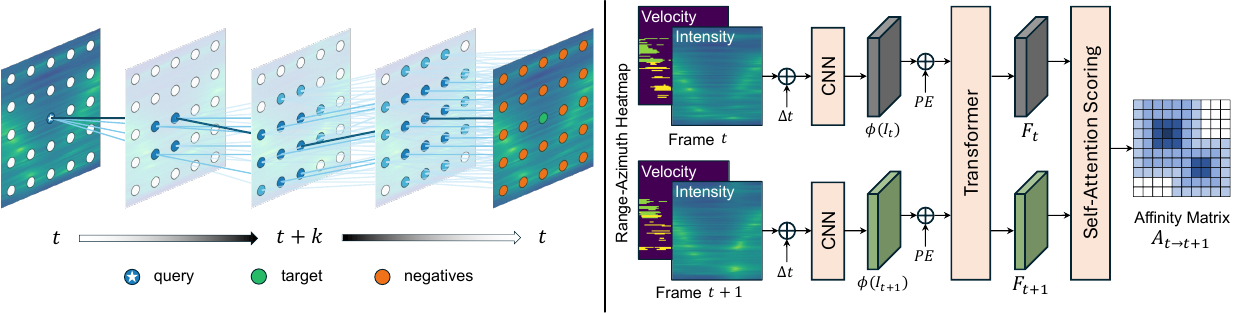}\vspace{-0.10in}
    \caption{Overview of the proposed cross-frame correspondence framework. 
    $\mathbf{Left}$:\,training with a palindrome random walk objective, which encourages a random walker to return to its starting node after a forward-backward traversal. 
    $\mathbf{Right}$:\,given two RA tensors, we construct a cross-frame graph over RA cells and apply a global matching architecture to estimate soft correspondence probabilities.}
    \label{fig:model}
\end{figure}

We formulate RF correspondence on the RA grid as a probabilistic walk on a space-time graph, 
where each RA cell is treated as a node and inter-frame edges encode soft correspondence probabilities. 
An overview is shown in Fig.~\ref{fig:model}.
Following GMRW~\citep{shrivastava2024self}, a CNN encoder and global matching transformer map two RF observations $\mathbf{X}_t$ and $\mathbf{X}_{t+1}$ to cell embeddings 
$\mathbf{F}_t, \mathbf{F}_{t+1} \in \mathbb{R}^{N \times d}$. 
We define the transition probability from source cell $i$ at time $t$ to target cell $j$ at time $t+1$ as
\begin{equation}
A_{t \rightarrow t+1}(i,j)
=
\frac{\exp\!\left(\mathbf{f}_{t,i}^{\top}\mathbf{f}_{t+1,j}/\tau\right)}
{\sum_{\ell=1}^{N}\exp\!\left(\mathbf{f}_{t,i}^{\top}\mathbf{f}_{t+1,\ell}/\tau\right)},
\label{eq:transition_matrix}
\end{equation}
where $\tau$ is a temperature parameter. 
Each row of $A_{t \rightarrow t+1}$ defines a probability distribution over target RA cells in frame $t+1$.
To capture longer-range temporal structure, we compose adjacent-frame transitions into a $k$-step walk
\begin{equation}
\bar{A}_{t \rightarrow t+k}
=
\prod_{s=0}^{k-1} A_{t+s \rightarrow t+s+1} .
\label{eq:composed_transition}
\end{equation}
Under this composition, transient clutter and unstable reflections tend to yield diffuse paths whose probability decays across steps, whereas persistent scene structures induce sharper paths across time. 
We adopt the palindrome training scheme of CRW~\citep{jabri2020space}, pairing a forward walk from $t$ to $t+k$ with its reverse walk. Reversing an RF sequence also requires inverting the sign of the Doppler channel, since radial velocity flips under time reversal.
We enforce cycle consistency by requiring a random walker to return to its starting node after the forward--backward traversal:
\begin{equation}
\mathcal{L}_{\mathrm{rw}}^{(k)}
=
\mathcal{L}_{\mathrm{CE}}
\!\left(
\bar{A}_{t \rightarrow t+k}\bar{A}_{t+k \rightarrow t},
I
\right),
\end{equation}
where $\mathcal{L}_{\mathrm{CE}}$ denotes row-wise cross-entropy and $I$ is the identity matrix. 
Cycle consistency admits a trivial solution in which the model simply copies absolute RA indices.
To prevent this, we randomly shift the 2D sinusoidal positional encoding during training, as in \cite{shrivastava2024self}.

\subsection{Doppler Consistency Supervision}
\label{method:doppler}

Pure cycle consistency does not guarantee physically correct correspondences. An incorrect transition may still participate in a self-consistent forward–backward walk if the reverse transition compensates for it. To address this limitation, we impose Doppler consistency directly on each one-step transition matrix, providing physics-grounded local supervision that requires the correspondence-induced radial displacement to agree with the Doppler-measured relative motion. This constraint reduces the feasible correspondence space by penalizing cycle-consistent but physically incompatible transitions. It is particularly beneficial in feature-poor or repetitive RF scenes, where intensity-based matching may otherwise admit multiple geometrically plausible correspondences.

\begin{figure}
    \centering
    \includegraphics[width=\linewidth]{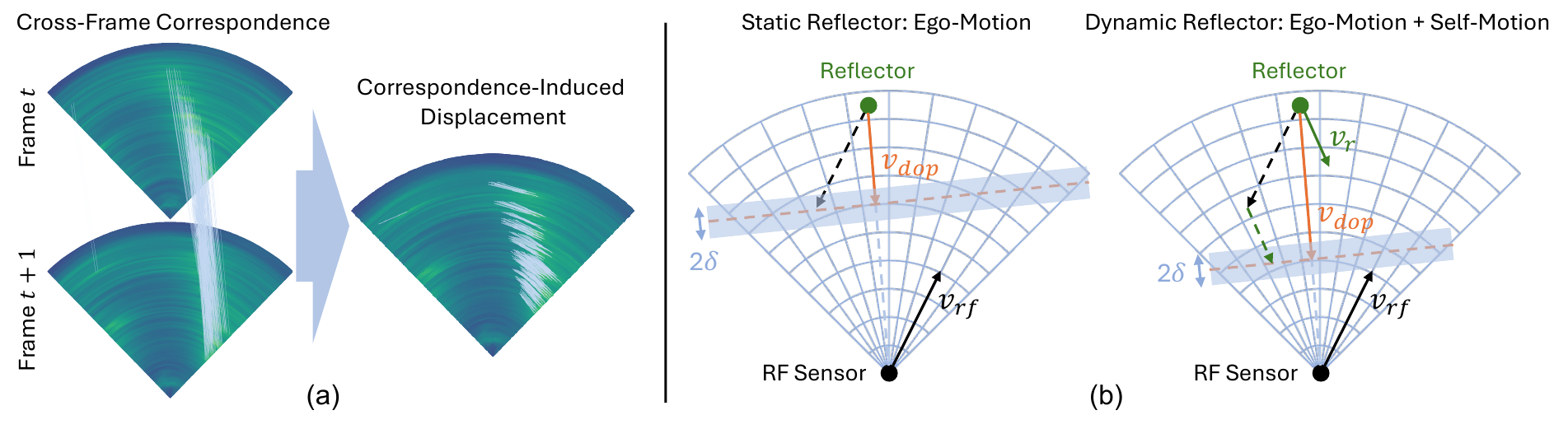}\vspace{-0.10in}
    \caption{Doppler provides a first-order physical constraint on correspondence-induced motion. 
    $\mathbf{(a)}$ Cross-frame correspondence induces a per-cell displacement field on the RA grid.
    $\mathbf{(b)}$ Doppler measures the sensor-reflector relative radial velocity, which arises from sensor ego-motion alone for a static reflector and from ego-motion together with the reflector's own motion for a dynamic one.}
    \label{fig:doppler}
\end{figure}

As illustrated in Fig.~\ref{fig:doppler}(a), the learned transition matrix $A_{t \rightarrow t+1}$ induces a soft displacement field over RA cells. For a source cell $i$, we define its expected target location and correspondence-induced displacement as
\begin{equation}
\tilde{\mathbf{p}}_i
=
\sum_{j=1}^{N}
A_{t \rightarrow t+1}(i,j)\,\mathbf{p}_j ,
\qquad
\boldsymbol{\delta}_i
=
\tilde{\mathbf{p}}_i - \mathbf{p}_i ,
\end{equation}
where $\mathbf{p}_i = [r_i \cos \theta_i \,, r_i \sin \theta_i]^\top$ is the Cartesian coordinate of the source cell $i$. 
Here, $\tilde{\mathbf{p}}_i$ denotes the expected target location corresponding to source cell $i$, not the location of target cell $i$. Let
\begin{equation}
\mathbf{u}_i
=
\frac{\mathbf{p}_i}{\|\mathbf{p}_i\|_2}
=
[\cos\theta_i \,,\, \sin\theta_i]^\top.
\end{equation}
denote the unit line-of-sight direction of that cell. Doppler measures the radial component of the relative velocity between the RF sensor and the observed reflector. Therefore, over a short inter-frame interval $\Delta t$, the radial projection of the correspondence-induced displacement should satisfy the first-order relation
\begin{equation}
\mathbf{u}_i^\top \boldsymbol{\delta}_i
\;\approx\;
\int_t^{t+\Delta t} - v_i^{\mathrm{dop}}(\tau)\, d\tau
\;\approx\;
- \Delta t \, v_i^{\mathrm{dop}} ,
\label{eq:doppler_consistency}
\end{equation}
where $v_i^{\mathrm{dop}}$ is the measured Doppler velocity at cell $i$ and $v_i^{\mathrm{dop}}>0$ indicates that the reflector approaches the sensor.
The relation in Eq.~\ref{eq:doppler_consistency} holds for both static and dynamic reflectors, as illustrated in Fig.~\ref{fig:doppler}(b). For a static reflector, the relative radial motion arises solely from sensor ego-motion, whereas for a dynamic reflector it additionally includes the reflector’s self-motion. Since both Doppler and correspondence-induced displacement measure the same sensor–reflector relative motion, the constraint remains valid in either case when the same physical scattering point is tracked.

Eq.~\ref{eq:doppler_consistency} further provides a geometric constraint on the target correspondence. For a given source cell $i$ and measured Doppler velocity $v_i^{\mathrm{dop}}$, all target locations satisfying Eq.~\ref{eq:doppler_consistency} lie on a line perpendicular to the line-of-sight direction $\mathbf{u}_i$, as illustrated in Fig.~\ref{fig:doppler}(b). Requiring exact equality is overly restrictive due to RA discretization and the first-order approximation over a finite $\Delta t$. We therefore introduce a tolerance margin $\delta>0$, which expands the ideal line into an admissible band of width $2\delta$, and penalize only deviations beyond this band. Accordingly, we define the Doppler consistency loss as
\begin{equation}
\mathcal{L}_{\mathrm{dop}}
=
\frac{1}{\sum_i q_i}
\sum_{i=1}^{N}
q_i\,
\max\!\left(
0,\,
\left|
\mathbf{u}_i^\top \boldsymbol{\delta}_i
+
\Delta t\, v_i^{\mathrm{dop}}
\right|
-
\delta
\right),
\label{eq:doppler_loss}
\end{equation}
where $q_i \in [0,1]$ is a data-dependent weight that downweights unreliable Doppler measurements.

\subsection{Training Objective}

\noindent\textbf{Group composition consistency.}
Long-horizon odometry is obtained by recursively composing pairwise motions, so local correspondence errors can accumulate into trajectory drift.
We therefore encourage the learned correspondences to be temporally compositional.

Given a temporal group $(\mathbf{X}_t, \mathbf{X}_{t+1}, \ldots, \mathbf{X}_{t+g})$, 
we construct the endpoint correspondence from $\mathbf{X}_t$ and $\mathbf{X}_{t+g}$ in two ways.
The composed transition $\bar{A}_{t \rightarrow t+g}$ aggregates the intermediate one-step transitions via Eq.~\ref{eq:composed_transition}, while the direct transition $A_{t \rightarrow t+g}$ is obtained by applying Eq.~\ref{eq:transition_matrix} to the endpoint pair.
Since both describe correspondence between the same frames, we enforce their agreement using a row-wise symmetric KL divergence:
\begin{equation}
\mathcal{L}_{\mathrm{gc}}^{(g)}
=
\frac{1}{2}
D_{\mathrm{KL}}^{\mathrm{row}}
\!\left(
\bar{A}_{t \rightarrow t+g}
\,\middle\|\,
A_{t \rightarrow t+g}
\right)
+
\frac{1}{2}
D_{\mathrm{KL}}^{\mathrm{row}}
\!\left(
A_{t \rightarrow t+g}
\,\middle\|\,
\bar{A}_{t \rightarrow t+g}
\right).
\end{equation}
This regularizes locally plausible but mutually inconsistent transitions and reduces drift under recursive pose composition.

\noindent\textbf{Overall objective.}
Combining the above terms, the final training objective is
\begin{equation}
\mathcal{L}
=
\lambda_{\mathrm{rw}}
\frac{1}{\left| \mathcal{K} \right|}
\sum_{k \in \mathcal{K}}
\mathcal{L}_{\mathrm{rw}}^{(k)}
+
\lambda_{\mathrm{dop}}
\mathcal{L}_{\mathrm{dop}}
+
\lambda_{\mathrm{gc}}
\frac{1}{\left| \mathcal{G} \right|}
\sum_{g \in \mathcal{G}}
\mathcal{L}_{\mathrm{gc}}^{(g)},
\label{eq:total_loss}
\end{equation}
where $\mathcal{K}$ denotes the set of palindrome lengths used for random-walk supervision, 
and $\mathcal{G}$ denotes the set of temporal group sizes used for composition consistency. 
$\lambda_{\mathrm{rw}}$, $\lambda_{\mathrm{dop}}$, and $\lambda_{\mathrm{gc}}$ are scalar weights. 
All three losses are self-supervised and require no ground-truth poses or correspondence annotations.

\subsection{Odometry Estimation}

The transition matrix $A_{t \rightarrow t+1}$ defines, for each source RA cell $i$, 
a discrete distribution over candidate target cells in frame $t+1$. 
Rather than collapsing this distribution with an $\arg\max$, 
we recover a soft Cartesian correspondence by taking its expectation over the target-cell coordinates:
\begin{equation}
\tilde{\mathbf{p}}_i
=
\sum_{j=1}^{N}
A_{t \rightarrow t+1}(i,j)\,\mathbf{p}_j,
\qquad
\mathbf{\Sigma}_i
=
\sum_{j=1}^{N}
A_{t \rightarrow t+1}(i,j)
(\mathbf{p}_j-\tilde{\mathbf{p}}_i)
(\mathbf{p}_j-\tilde{\mathbf{p}}_i)^\top .
\label{eq:soft_corr}
\end{equation}
Here, $\tilde{\mathbf{p}}_i$ is the expected target location of source cell $i$, 
while $\mathbf{\Sigma}_i$ measures the spatial dispersion of its correspondence distribution. 
Not all RA cells should contribute equally to ego-motion estimation. 
Low-SNR bins and multipath reflections often induce diffuse transition distributions and should contribute less to the final motion estimate.
We therefore assign each source cell a match-quality prior
\begin{equation}
\bar{w}_i
=
m_i\,
\exp\!\left(-\beta\,\mathrm{tr}(\mathbf{\Sigma}_i)\right)
\left(
1-\frac{\mathcal{H}(A_i)}{\log N}
\right),
\qquad
\mathcal{H}(A_i)
=
-\sum_{j=1}^{N}
A_{t \rightarrow t+1}(i,j)\log A_{t \rightarrow t+1}(i,j),
\label{eq:quality_weight}
\end{equation}
where $A_i$ denotes the $i$-th row of $A_{t \rightarrow t+1}$, 
$\mathcal{H}(A_i)$ is its entropy, 
and $m_i \in [0,1]$ is a measurement-confidence prior computed from the source RA response. 
This weight favors source cells with strong measurements and sharp, spatially concentrated correspondence distributions.

A more fundamental challenge is that only the static part of the scene should obey a single global $\mathrm{SE}(2)$ motion. 
Reflectors on walking people, opening doors, or other independently moving objects violate this assumption and should be treated as outliers rather than forced into the rigid fit. 
To this end, we estimate odometry with a robust M-estimator:
\begin{equation}
(\mathbf{R}_t,\mathbf{t}_t)
=
\arg\min_{\mathbf{R}\in \mathrm{SO}(2),\,\mathbf{t}\in\mathbb{R}^2}
\sum_{i=1}^{N}
\bar{w}_i\,
\rho\!\left(
\left\|
\mathbf{R}\mathbf{p}_i + \mathbf{t} - \tilde{\mathbf{p}}_i
\right\|_2
\right),
\label{eq:robust_se2_fit}
\end{equation}
where $\rho(\cdot)$ is the Huber loss,
\begin{equation}
\rho(r)=
\begin{cases}
\frac{1}{2}r^2, & r \le c,\\[2pt]
c\left(r-\frac{1}{2}c\right), & r > c,
\end{cases}
\end{equation}
with threshold $c > 0$.
Eq.~\ref{eq:robust_se2_fit} preserves high-confidence static correspondences while suppressing large-residual outliers caused by dynamic objects or gross matching errors. We solve it via iteratively reweighted least squares and provide the complete closed-form derivation in Appendix~\ref{app:solver}.

\section{Experiments}

\subsection{Experimental Setup}

\noindent\textbf{Datasets.}
We evaluate on two complementary datasets that differ substantially in carrier frequency, bandwidth, antenna configuration, sensing range, and scene dynamics.

\emph{Radarize dataset.}
Radarize~\cite{sie2024radarize} is a public dataset collected with a TI IWR1843 single-chip mmWave radar operating in the 76\textendash81\,GHz band.
The radar has 3 transmit and 4 receive antennas and provides a maximum range of 4.284\,m, a range resolution of 4\,cm, and a frame rate of 30\,Hz.
The dataset contains 146 trajectories recorded across 4 buildings, covering approximately 4.7\,km in total.

\begin{wrapfigure}{r}{0.48\columnwidth}
    \vspace{0.4\baselineskip}
    \centering
    \includegraphics[width=0.46\columnwidth]{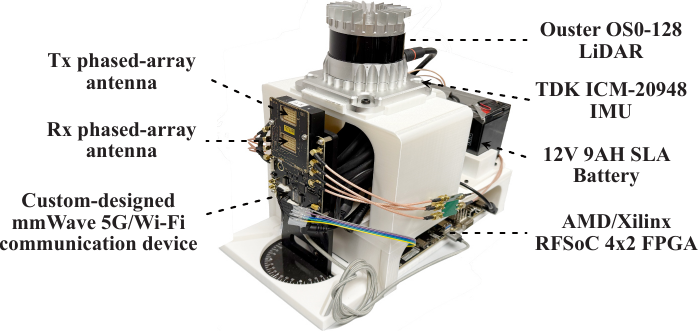}
    \caption{Custom-designed 60\,GHz mmWave monostatic ISAC platform.}
    \label{fig:dataset}
\end{wrapfigure}

\emph{\pname dataset.}
We collect our own data with a custom-designed 60\,GHz monostatic integrated sensing and communication (ISAC) platform (Fig.~\ref{fig:dataset}).
The platform simultaneously transmits and receives OFDM communication waveforms, from which RA frames are formed.
It has a bandwidth of 1.2288\,GHz, corresponding to a range resolution of about 12\,cm, uses 16 transmit and 16 receive phased-array antennas, and records frames at 10\,Hz.
The \pname dataset contains 80 trajectories with a total length of 3.2\,km.
In contrast to Radarize, it is recorded in the presence of moving pedestrians, which introduce substantial dynamic clutter into the RF observations.

\noindent\textbf{Evaluation metrics.}
Following the protocol of Radarize~\cite{sie2024radarize}, we adopt two standard trajectory metrics.
Absolute Trajectory Error (ATE) measures the global consistency of the estimated trajectory, whereas Relative Error (RE) measures local odometry accuracy as the drift accumulated over short trajectory segments.
We report both metrics separately for translation (m) and rotation (deg) and compute them with the \texttt{evo} toolbox~\cite{grupp2017evo}.

\noindent\textbf{Baselines.}
We compare against five baselines that span the major paradigms of RF odometry.
The first two are classical scan-matching methods that require no training.
(i)~ICP~\cite{besl1992method} iteratively registers consecutive RF scans by minimizing point-to-point distances.
(ii)~NDT~\cite{biber2003normal} models each scan as a set of local normal distributions and registers consecutive scans by maximizing their likelihood.
The remaining three are learning-based methods.
(iii)~RadarHD~\cite{prabhakara2023high} upsamples range-azimuth heatmaps with a U-Net and feeds the enhanced heatmaps to a scan-matching SLAM backend.
(iv)~milliEgo~\cite{lu2020milliego} is a learning-based fusion approach that combines mmWave radar with an inertial measurement unit (IMU).
(v)~Radarize~\cite{sie2024radarize} decomposes ego-motion into Doppler-based translation estimation and correlation-based rotation estimation.
All three learning-based baselines are trained with ground-truth pose supervision, whereas \pname uses none.
All methods are evaluated on the same test trajectories.

\noindent\textbf{Implementation details.}
The CNN encoder consists of two convolutional blocks followed by a linear projection that maps each RA cell to a $d{=}128$-dimensional embedding.
The transformer has 4 layers with 8 attention heads, and the temperature of the transition matrix is set to $\tau{=}\sqrt{d}$.
During training, we use palindrome walk lengths $\mathcal{K}{=}\{1,2,3\}$ and group composition sizes $\mathcal{G}{=}\{2,3\}$.
The loss weights are $\lambda_{\mathrm{rw}}{=}1.0$, $\lambda_{\mathrm{dop}}{=}0.5$, and $\lambda_{\mathrm{gc}}{=}0.2$, and the tolerance margin in the Doppler consistency loss is $\delta{=}0.02$\,m.
For odometry estimation, we set the dispersion decay factor to $\beta{=}5.0$ and the Huber threshold to $c{=}0.1$\,m.
All models are trained on 4 NVIDIA RTX A6000 GPUs.

\begin{table}[t]
\footnotesize
\centering
\setlength{\tabcolsep}{2pt}
\setlength{\aboverulesep}{0.4pt}
\setlength{\belowrulesep}{1.8pt}
\caption{Odometry accuracy on the Radarize (static) and \pname (dynamic) datasets.
ATE measures global trajectory consistency and RE measures local odometry drift, each reported for translation (Pos, m) and rotation (Rot, deg).
Lower is better for all metrics, and the best result in each column is shown in bold.
Supervision indicates whether a method requires ground-truth poses for training.
}
\begin{tabular}{
m{0.14\linewidth}<{\raggedright}
m{0.08\linewidth}<{\centering}
m{0.08\linewidth}<{\centering}
m{0.08\linewidth}<{\centering}
m{0.08\linewidth}<{\centering} 
m{0.08\linewidth}<{\centering} 
m{0.08\linewidth}<{\centering} 
m{0.08\linewidth}<{\centering}
m{0.08\linewidth}<{\centering}
m{0.08\linewidth}<{\centering}}
\toprule
\multirow{3}{*}{\vspace{-0.08in} Methods} & \multirow{3}{*}{\vspace{-0.08in} \!\!\!\!\!\!\!Supervision\!\!\!\!\!} & \multicolumn{4}{c}{Radarize\,(Static Scenes)} & \multicolumn{4}{c}{\pname\,(Dynamic Scenes)} \\
\cmidrule(lr){3-6}\cmidrule(lr){7-10}
& & \multicolumn{2}{c}{ATE} & \multicolumn{2}{c}{RE} & \multicolumn{2}{c}{ATE} & \multicolumn{2}{c}{RE} \\
\cmidrule(lr){3-4}\cmidrule(lr){5-6}\cmidrule(lr){7-8}\cmidrule(lr){9-10}
& & Pos (m) & Rot (deg) & Pos (m) & Rot (deg) & Pos (m) & Rot (deg) & Pos (m) & Rot (deg) \\
\midrule
ICP \cite{besl1992method}           & \xmark & 8.613 & 7.104 & 0.027 & 0.022 & 16.482 & 8.036 & 0.052 & 0.035 \\
NDT \cite{biber2003normal}          & \xmark & 7.421 & 6.372 & 0.023 & 0.019 & 14.208 & 7.184 & 0.045 & 0.031 \\
\midrule
RadarHD \cite{prabhakara2023high}   & \cmark & 6.367 & 5.820 & 0.020 & 0.016 & 12.724 & 6.215 & 0.041 & 0.027 \\
milliEgo \cite{lu2020milliego}      & \cmark & 4.732 & 7.774 & 0.014 & 0.020 & 7.406 & 5.123 & 0.022 & 0.024 \\
Radarize \cite{sie2024radarize}     & \cmark & \textbf{1.657} & \textbf{2.445} & \textbf{0.007} & \textbf{0.008} & 5.642 & 4.938 & 0.018 & 0.022 \\
\rowcolor[HTML]{D7D9E3}
\pname(Ours)                        & \xmark & 1.934 & 3.191 & 0.008 & 0.010 & \textbf{2.357} & \textbf{2.607} & \textbf{0.010} & \textbf{0.013} \\
\bottomrule
\end{tabular}
\label{table:quantitative-performance}
\end{table}

\subsection{Main Results}

For the Radarize dataset, we follow the official train/test split~\cite{sie2024radarize}.
For the \pname dataset, we randomly assign 60\% of the trajectories to training and the remaining 40\% to testing.
The learning-based baselines are trained with ground-truth poses obtained from LiDAR and IMU, whereas \pname is trained only on unlabeled RF sequences.
Tab.~\ref{table:quantitative-performance} summarizes the quantitative results.

\begin{figure}[t]
    \centering
    \includegraphics[width=0.95\linewidth]{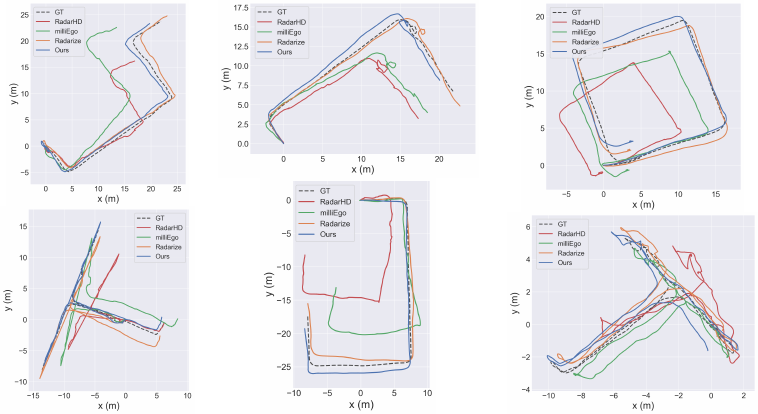}
    \caption{Qualitative trajectory comparison on the Radarize dataset. 
    Each subplot shows the estimated trajectory of all methods overlaid with the ground truth (GT) marked by the dashed black line.}
    \label{fig:radarize_traj}
\end{figure}

\noindent\textbf{Performance on static scenes.}
On the Radarize dataset, the supervised Radarize baseline achieves the best accuracy on all four metrics, as its Doppler-based translation estimator provides a strong inductive bias when the scene is static.
Without any pose supervision, \pname remains close to Radarize, and it outperforms all other baselines on every metric.
The training-free ICP and NDT perform worst, since coarse angular resolution, multipath, and sparse returns make direct geometric registration of RF scans unreliable.
These results show that self-supervised correspondence learning alone recovers ego-motion with accuracy approaching that of the strongest supervised method in static environments, as also reflected in the trajectories in Fig.~\ref{fig:radarize_traj}.

\begin{figure}[t]
    \centering
    \includegraphics[width=0.95\linewidth]{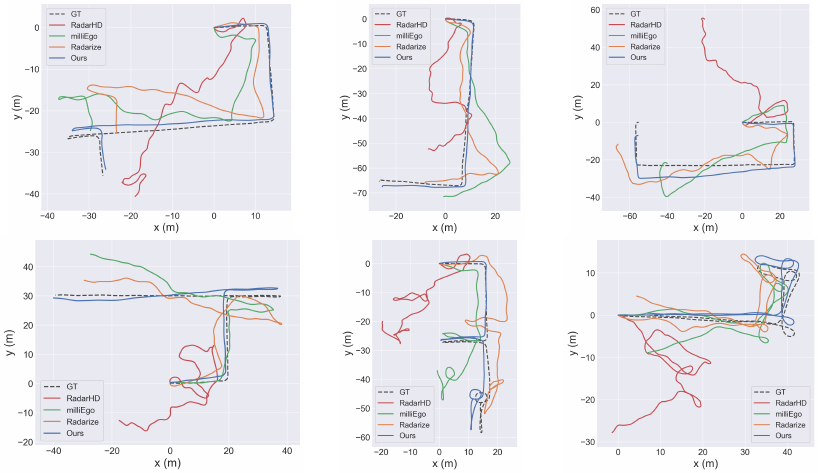}
    \caption{Qualitative trajectory comparison on the \pname dataset with dynamic pedestrians. 
    All supervised baselines exhibit notable degradation in the presence of moving objects.}
    \label{fig:isac_traj}
\end{figure}

\noindent\textbf{Performance on dynamic scenes.}
On the \pname dataset with frequent pedestrian motion, \pname achieves the best accuracy on all four metrics.
ICP and NDT yield the largest errors, because both assume a rigid scene and therefore register moving pedestrians as if they were static structure.
We attribute the robustness of \pname to two complementary mechanisms.
First, the cycle-consistent random walk objective suppresses correspondences on transient clutter, since moving objects induce diffuse paths whose probability decays under multi-step composition.
Second, the robust M-estimator in Eq.~\ref{eq:robust_se2_fit} downweights the remaining large-residual correspondences on independently moving objects.
Together, these mechanisms allow \pname to handle dynamic objects without explicit motion segmentation, as illustrated in Fig.~\ref{fig:isac_traj}.

\subsection{Ablation Study}

We further analyze the robustness of \pname to scene dynamics and the contribution of its key design choices.
All experiments in this section are performed on the \pname dataset, which provides a more challenging testbed due to scene dynamics and geometric degeneracy.

\begin{table}[t]
\footnotesize
\centering
\setlength{\aboverulesep}{0.4pt}
\setlength{\belowrulesep}{1.8pt}
\begin{minipage}[t]{0.46\linewidth}
\centering
\caption{ATE Pos (m) under increasing levels of scene dynamics.}
\setlength{\tabcolsep}{2pt}
\label{table:ablation-dynamics}
\begin{tabular}{
m{0.28\linewidth}<{\raggedright}
m{0.20\linewidth}<{\centering}
m{0.20\linewidth}<{\centering}
m{0.20\linewidth}<{\centering}}
\toprule
Methods & Low & Medium & High \\
\midrule
RadarHD \cite{prabhakara2023high} & 10.243 & 13.175 & 18.478 \\
milliEgo \cite{lu2020milliego}    & 6.812 & 7.934 & 9.105 \\
Radarize \cite{sie2024radarize}   & 4.831 & 6.042 & 8.813 \\
\rowcolor[HTML]{D7D9E3}
\pname                            & \textbf{2.285} & \textbf{2.672} & \textbf{3.214} \\
\bottomrule
\end{tabular}
\begin{tablenotes}[flushleft]
\footnotesize
\item[] Low, Medium, and High correspond to scenes with 0--2, 3--4, and 5+ moving pedestrians, respectively.
\end{tablenotes}
\end{minipage}
\hfill
\begin{minipage}[t]{0.52\linewidth}
\centering
\caption{Effect of group composition sizes $\mathcal{G}$. $\emptyset$ denotes training without $\mathcal{L}_{\mathrm{gc}}$.}
\setlength{\tabcolsep}{2pt}
\begin{tabular}{
m{0.19\linewidth}<{\centering}
m{0.14\linewidth}<{\centering}
m{0.18\linewidth}<{\centering}
m{0.14\linewidth}<{\centering}
m{0.18\linewidth}<{\centering}}
\toprule
\multirow{2}{*}{\vspace{-0.07in}$\mathcal{G}$} & \multicolumn{2}{c}{ATE} & \multicolumn{2}{c}{RE} \\
\cmidrule(lr){2-3}\cmidrule(lr){4-5}
& Pos (m) & Rot (deg) & Pos (m) & Rot (deg) \\
\midrule
$\emptyset$       & 2.675 & \textbf{2.487} & 0.013 & \textbf{0.011} \\
$\{2\}$           & 2.402 & 2.575 & 0.011 & 0.013 \\
\rowcolor[HTML]{D7D9E3}
$\{2,3\}$         & \textbf{2.357} & 2.607 & \textbf{0.010} & 0.013 \\
$\{2,3,4\}$       & 2.728 & 3.253 & 0.012 & 0.016 \\
\bottomrule
\end{tabular}
\label{table:ablation-group-size}
\end{minipage}
\end{table}

\noindent\textbf{Robustness to dynamic objects.}
To examine how scene dynamics affect each method, we partition the test trajectories into three groups by the number of moving pedestrians and report the ATE Pos of each group in Tab.~\ref{table:ablation-dynamics}.
\pname achieves the lowest error at every level, and its error increases by only 0.93\,m from the Low to the High group, compared with 2.29\,m to 8.24\,m for the supervised baselines.
RadarHD degrades the most, since its scan-matching backend assumes a rigid scene and is therefore directly corrupted by moving pedestrians.
milliEgo degrades more gracefully owing to its IMU branch, yet its fusion network still ingests corrupted RF features and cannot fully compensate.
Radarize attains lower errors than milliEgo at every level, but its error grows faster, from 4.83\,m to 8.81\,m, because its Doppler-based translation estimator conflates pedestrian velocities with ego-motion.

\noindent\textbf{Effect of group composition size.}
Tab.~\ref{table:ablation-group-size} varies the temporal group sizes $\mathcal{G}$ in the group composition consistency loss.
Without this loss ($\mathcal{G}{=}\emptyset$), the palindrome objective constrains each forward walk only through its own backward return, so locally plausible but mutually incompatible transitions can still accumulate into drift.
Adding $\mathcal{G}{=}\{2\}$ and then $\{2,3\}$ steadily improves translational accuracy, and $\{2,3\}$ reduces ATE Pos by 12\% and RE Pos by 23\% relative to $\emptyset$.
Extending the groups to $\{2,3,4\}$ degrades all metrics, as direct transitions over longer horizons become less reliable and enforcing agreement with them propagates their errors back to the one-step transitions.
We also observe a mild trade-off in rotation, where $\mathcal{L}_{\mathrm{gc}}$ slightly increases ATE Rot from 2.487\,deg to 2.607\,deg.
Since translational drift dominates the overall trajectory error, we adopt $\mathcal{G}{=}\{2,3\}$ by default.

\begin{wrapfigure}{r}{0.60\columnwidth}
\vspace{-0.5\baselineskip}
\centering
\begin{minipage}[b]{0.49\linewidth}
    \centering
    \includegraphics[width=\linewidth]{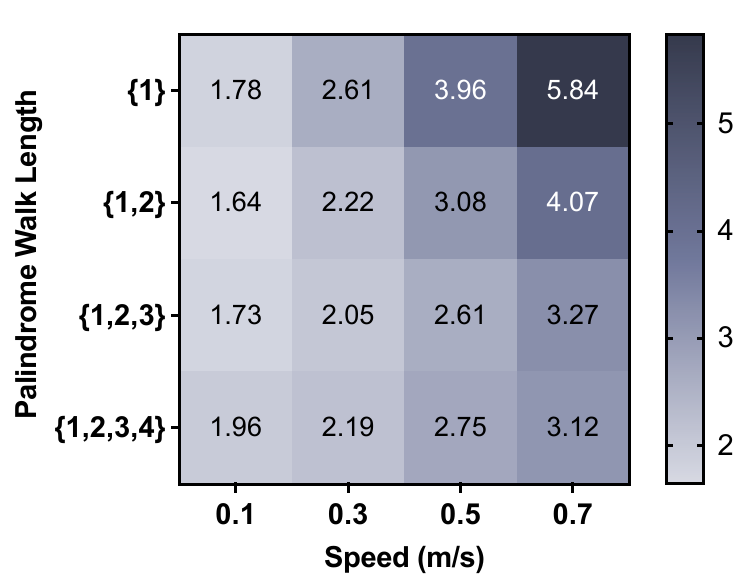}\\
    {\footnotesize (a) Varying speed.}
\end{minipage}
\hfill
\begin{minipage}[b]{0.49\linewidth}
    \centering
    \includegraphics[width=\linewidth]{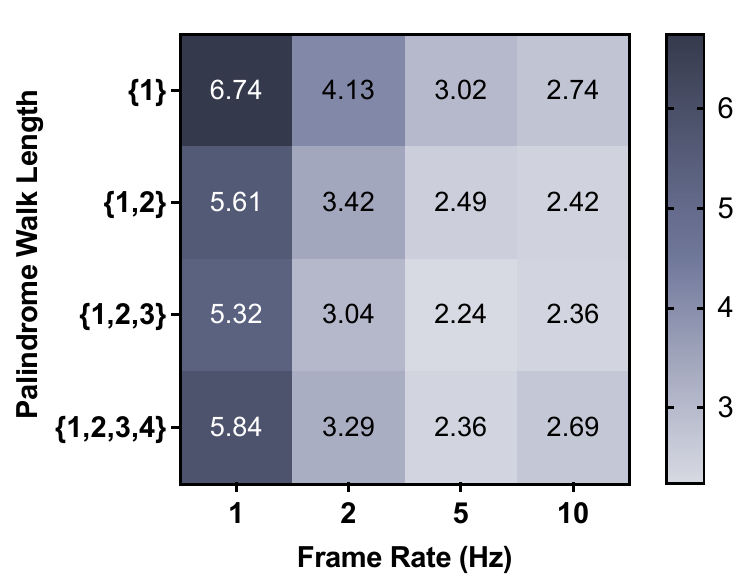}\\
    {\footnotesize (b) Varying frame rate.}
\end{minipage}
\caption{Effect of palindrome walk length $\mathcal{K}$ across speeds (m/s) and frame rates (Hz), measured by ATE Pos (m).}
\label{fig:ablation-walk-length}
\vspace{-0.5\baselineskip}
\end{wrapfigure}

\noindent\textbf{Effect of palindrome walk length.}
The palindrome walk length $\mathcal{K}$ controls the temporal horizon over which cycle consistency is enforced, and its optimal choice depends on the effective inter-frame motion. 
We disentangle this dependence along two complementary axes in Fig.~\ref{fig:ablation-walk-length}.

Fig.~\ref{fig:ablation-walk-length}(a) varies the test-time platform speed while training each model with a different $\mathcal{K}$. 
The penalty of longer chains is most pronounced at slow speeds, where per-frame displacements are small and multi-step composition mainly amplifies correspondence noise. 
As speed grows, the gap between walk lengths widens substantially, since longer palindromes provide stronger multi-step regularization when inter-frame displacement is large.
Fig.~\ref{fig:ablation-walk-length}(b) jointly varies the frame rate at training and inference. 
At 1\,Hz, sparse temporal sampling produces large per-step motions, and walks beyond two steps drift outside the reliable correspondence horizon. 
As the frame rate increases, the effective per-step displacement shrinks and longer chains become beneficial, with $\mathcal{K}{=}\{1,2,3\}$ attaining the lowest error.

\begin{wrapfigure}{r}{0.38\columnwidth}
\centering
\includegraphics[width=\linewidth]{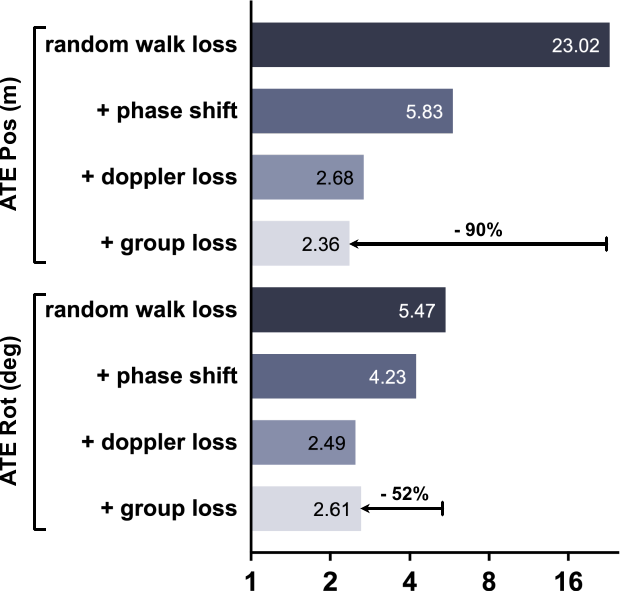}\vspace{-0.10in}
\caption{Cumulative ablation of training components. Each row adds one component to the row above.}
\label{fig:ablation-components}
\end{wrapfigure}

\noindent\textbf{Effect of individual components.}
To understand where the gains come from, we progressively add each component to the base random walk objective $\mathcal{L}_{\mathrm{rw}}$ and report the results in Fig.~\ref{fig:ablation-components}.
Together, these components reduce ATE Pos by 90\% and ATE Rot by 52\%.

The random phase shift yields the largest translational gain, reducing ATE Pos from 23.02\,m to 5.83\,m.
Without it, the positional encodings of consecutive frames are aligned, so cycle consistency can be trivially satisfied by copying RA indices and the transition matrix collapses toward the identity.
The phase shift removes this shortcut and forces the model to match cells by their RF appearance.
The Doppler consistency loss $\mathcal{L}_{\mathrm{dop}}$ yields the largest rotational gain, reducing ATE Rot from 4.23\,deg to 2.49\,deg, and also lowers ATE Pos by 54\%.
In feature-poor scenes, translational ambiguity in the correspondences couples into rotation during $\mathrm{SE}(2)$ fitting, and anchoring radial displacements to measured Doppler velocities resolves this ambiguity.
Finally, the group composition consistency loss $\mathcal{L}_{\mathrm{gc}}$ further reduces ATE Pos to 2.36\,m by mitigating drift under recursive pose composition.

\section{Related Work}

\subsection{Self-Supervised Space-Time Correspondence}

Learning correspondences across frames underlies video segmentation, tracking, and 3D reconstruction~\citep{perazzi2016benchmark, schonberger2016structure}.
Supervised optical flow~\citep{dosovitskiy2015flownet, sun2018pwc, teed2020raft} and tracking-any-point (TAP) methods~\citep{doersch2022tap, harley2022particle, doersch2023tapir, karaev2024cotracker, karaev2025cotracker3} achieve high accuracy on natural videos but rely largely on synthetic annotations or pseudo-labels.
Self-supervised methods instead learn correspondences from unlabeled videos through proxy tasks~\citep{vondrick2018tracking, lai2020mast}, or through temporal cycle consistency~\citep{wang2019learning, li2019joint}.
CRW~\citep{jabri2020space} formulates correspondence as a contrastive random walk on a space-time graph.
This formulation has been extended to multiscale pixel trajectories~\citep{bian2022learning}, combined with a global matching transformer for all-pairs any-point tracking in GMRW~\citep{shrivastava2024self}, and applied to cross-modal matching among RGB, depth, and thermal images~\citep{shrivastava2025self}.

More fundamentally, cycle consistency is derived from the observations alone and is satisfied by any set of transitions whose composition returns to the start.
In natural images, rich appearance makes it a reliable proxy for correct matching, but RA heatmaps have coarse resolution and repetitive structures that admit many cycle-consistent yet physically incorrect matches.
\pname brings the random-walk formulation to RF and imposes Doppler consistency directly on its soft transition probabilities rather than on regressed flow, so that every correspondence between dense RA cells must agree with the radial motion measured by the radar.

\subsection{RF-Based Odometry}

Geometric radar odometry estimates motion by aligning consecutive RF observations.
Early methods extract and match landmarks from scanning radar~\citep{cen2018precise}.
CFEAR~\citep{adolfsson2021cfear} registers sparse oriented surface points, and RadarHD~\citep{prabhakara2023high} learns to generate LiDAR-like point clouds from single-chip mmWave radar for subsequent scan matching.
However, coarse angular resolution, multipath ghosts, and geometric degeneracy in feature-poor environments make such registration unreliable, especially for low-cost single-chip radars.
Radar-inertial systems improve motion observability by fusing an IMU through filtering or optimization~\citep{kramer2020radar, doer2020ekf, zhuang20234d} or through deep sensor fusion~\citep{lu2020milliego}.
These systems, however, require additional sensing, calibration, and synchronization.
Learning-based radar odometry has largely relied on pose supervision.
Recent mmWave pipelines~\citep{prabhakara2023high, sie2024radarize} require co-located LiDAR or ground-truth trajectories during training.

Doppler provides a complementary motion cue by measuring the radial velocity of each reflector.
Radarize~\citep{sie2024radarize} combines Doppler-based translation with correlation-based rotation estimation, while PanoRadar~\citep{lai2024enabling} and CartoRadar~\citep{lai2025rf} incorporate Doppler into motion compensation and RF SLAM.
These methods treat Doppler as a measurement of ego-motion and therefore assume that reflectors are static.
\pname instead uses Doppler as local supervision for correspondence learning, and recovers ego-motion through robust $\mathrm{SE}(2)$ registration that suppresses independently moving correspondences.
The resulting pipeline requires neither ground-truth trajectories nor auxiliary sensors.

\section{Conclusion}

We presented \pname, a self-supervised, RF-only odometry framework that learns cross-frame correspondences without ground-truth poses or auxiliary sensors.
By combining cycle-consistent learning with Doppler-based physical supervision, \pname achieves competitive accuracy in static environments and substantial improvements in dynamic scenes.

\noindent\textbf{Limitations and future work.}
Several limitations remain.
First, global correspondence between RA frames scales quadratically with the number of RA cells $N$.
Second, the current formulation estimates only planar $\mathrm{SE}(2)$ motion.
Third, each model is trained and evaluated on a single radar configuration, and zero-shot transfer across RF sensors with different carrier frequencies, bandwidths, and antenna layouts remains unexplored.
However, the label-free training formulation provides a natural path toward addressing this limitation.
Since \pname requires no pose labels and represents RF observations using a common intensity and radial-velocity interface across radar configurations, it can in principle be trained on unlabeled sequences pooled from heterogeneous RF platforms.
Future work will therefore explore efficient correspondence architectures, full $\mathrm{SE}(3)$ motion estimation, and large-scale self-supervised pretraining of general-purpose RF motion representations across sensors.

\newpage
\section*{Acknowledgments}
We sincerely thank the reviewers for their insightful comments and suggestions.
This work is supported in part by the National Science Foundation (NSF) under Award No.~ECCS-2434001.

{
    \small
    \bibliographystyle{ACM-Reference-Format}
    \bibliography{reference}
}

\appendix

\section{Solving the Robust SE(2) Registration}
\label{app:solver}

In the main text, Eq.~\ref{eq:robust_se2_fit} formulates odometry estimation as a robust weighted registration problem on $\mathrm{SE}(2)$.
Here we provide the complete derivation of its solution.

Recall that the optimization problem is
\begin{equation}
(\mathbf{R}^*,\mathbf{t}^*)
=
\arg\min_{\mathbf{R}\in \mathrm{SO}(2),\,\mathbf{t}\in\mathbb{R}^2}
\sum_{i=1}^{N}
\bar{w}_i\,
\rho\!\left(
\left\|
\mathbf{R}\mathbf{p}_i + \mathbf{t} - \tilde{\mathbf{p}}_i
\right\|_2
\right),
\tag{A.1}
\label{eq:app_robust}
\end{equation}
where $\mathbf{p}_i \in \mathbb{R}^2$ is the Cartesian position of source RA cell $i$,
$\tilde{\mathbf{p}}_i \in \mathbb{R}^2$ is the expected target position defined in Eq.~\ref{eq:soft_corr},
$\bar{w}_i \ge 0$ is the match-quality prior from Eq.~\ref{eq:quality_weight},
and $\rho(\cdot)$ is the Huber loss with threshold $c > 0$.

\subsection*{A.1\quad Iteratively Reweighted Least Squares}

Because the Huber loss is non-quadratic, we solve Eq.~\ref{eq:app_robust} via iteratively reweighted least squares (IRLS).
The key observation is that the Huber loss admits a \emph{half-quadratic} interpretation.
For any residual $r_i = \left\| \mathbf{R}\mathbf{p}_i + \mathbf{t} - \tilde{\mathbf{p}}_i \right\|_2$, the first derivative of $\rho$ is
\begin{equation}
\rho'(r)
=
\begin{cases}
r, & r \le c, \\[3pt]
c, & r > c.
\end{cases}
\tag{A.2}
\end{equation}
The IRLS weight function, defined as the ratio $\psi(r) = \rho'(r)/r$ for $r > 0$, is
\begin{equation}
\psi(r)
=
\begin{cases}
1, & r \le c, \\[3pt]
c / r, & r > c.
\end{cases}
\tag{A.3}
\label{eq:irls_weight}
\end{equation}
At each IRLS iteration, the effective weight for cell $i$ becomes
\begin{equation}
\omega_i = \bar{w}_i \cdot \psi(r_i),
\tag{A.4}
\label{eq:effective_weight}
\end{equation}
and the original problem reduces to a \emph{weighted least squares} rigid registration
\begin{equation}
(\mathbf{R}^*,\mathbf{t}^*)
=
\arg\min_{\mathbf{R}\in \mathrm{SO}(2),\,\mathbf{t}\in\mathbb{R}^2}
\sum_{i=1}^{N}
\omega_i\,
\left\|
\mathbf{R}\mathbf{p}_i + \mathbf{t} - \tilde{\mathbf{p}}_i
\right\|_2^2.
\tag{A.5}
\label{eq:wls}
\end{equation}
This subproblem has a closed-form solution, which we derive below.

\subsection*{A.2\quad Closed-Form Weighted SE(2) Registration}

We first eliminate the translational variable by centering both point sets.
Define the weighted centroids
\begin{equation}
\boldsymbol{\mu}_{\mathbf{p}}
=
\frac{\sum_{i=1}^{N} \omega_i \, \mathbf{p}_i}{\sum_{i=1}^{N} \omega_i},
\qquad
\boldsymbol{\mu}_{\tilde{\mathbf{p}}}
=
\frac{\sum_{i=1}^{N} \omega_i \, \tilde{\mathbf{p}}_i}{\sum_{i=1}^{N} \omega_i},
\tag{A.6}
\label{eq:centroids}
\end{equation}
and the centered coordinates
\begin{equation}
\hat{\mathbf{p}}_i = \mathbf{p}_i - \boldsymbol{\mu}_{\mathbf{p}},
\qquad
\hat{\tilde{\mathbf{p}}}_i = \tilde{\mathbf{p}}_i - \boldsymbol{\mu}_{\tilde{\mathbf{p}}}.
\tag{A.7}
\end{equation}

Substituting $\mathbf{t} = \boldsymbol{\mu}_{\tilde{\mathbf{p}}} - \mathbf{R}\boldsymbol{\mu}_{\mathbf{p}} + \boldsymbol{\epsilon}$ into Eq.~\ref{eq:wls} and expanding, the objective decouples into a term depending only on $\mathbf{R}$ (through centered coordinates) and a term depending only on $\boldsymbol{\epsilon}$:
\begin{equation}
\sum_{i=1}^{N} \omega_i
\left\|
\mathbf{R}\hat{\mathbf{p}}_i - \hat{\tilde{\mathbf{p}}}_i + \boldsymbol{\epsilon}
\right\|_2^2
=
\sum_{i=1}^{N} \omega_i
\left\|
\mathbf{R}\hat{\mathbf{p}}_i - \hat{\tilde{\mathbf{p}}}_i
\right\|_2^2
+
\left(\sum_{i=1}^{N} \omega_i\right)
\left\| \boldsymbol{\epsilon} \right\|_2^2,
\tag{A.8}
\end{equation}
where the cross term vanishes because $\sum_i \omega_i \hat{\mathbf{p}}_i = \sum_i \omega_i \hat{\tilde{\mathbf{p}}}_i = \mathbf{0}$ by definition of the centroids.
Therefore $\boldsymbol{\epsilon}^* = \mathbf{0}$, and the optimal translation is
\begin{equation}
\mathbf{t}^* = \boldsymbol{\mu}_{\tilde{\mathbf{p}}} - \mathbf{R}^* \boldsymbol{\mu}_{\mathbf{p}}.
\tag{A.9}
\label{eq:opt_t}
\end{equation}

\noindent\textbf{Optimal rotation.}
It remains to solve
\begin{equation}
\mathbf{R}^*
=
\arg\min_{\mathbf{R}\in \mathrm{SO}(2)}
\sum_{i=1}^{N}
\omega_i
\left\|
\mathbf{R}\hat{\mathbf{p}}_i - \hat{\tilde{\mathbf{p}}}_i
\right\|_2^2.
\tag{A.10}
\label{eq:rot_only}
\end{equation}
Expanding the squared norm and dropping terms independent of $\mathbf{R}$, we obtain the equivalent maximization
\begin{equation}
\mathbf{R}^*
=
\arg\max_{\mathbf{R}\in \mathrm{SO}(2)}
\sum_{i=1}^{N}
\omega_i \,
\hat{\tilde{\mathbf{p}}}_i^\top
\mathbf{R}\hat{\mathbf{p}}_i
=
\arg\max_{\mathbf{R}\in \mathrm{SO}(2)}
\operatorname{tr}\!\left(
\mathbf{R}\mathbf{H}^\top
\right),
\tag{A.11}
\label{eq:trace_max}
\end{equation}
where $\mathbf{H} \in \mathbb{R}^{2 \times 2}$ is the weighted cross-covariance matrix
\begin{equation}
\mathbf{H}
=
\sum_{i=1}^{N}
\omega_i \,
\hat{\tilde{\mathbf{p}}}_i \,
\hat{\mathbf{p}}_i^\top.
\tag{A.12}
\label{eq:cross_cov}
\end{equation}

In two dimensions, $\mathbf{R}$ is parameterized by a single angle $\Delta\psi$,
\begin{equation}
\mathbf{R}(\Delta\psi)
=
\begin{bmatrix}
\cos\Delta\psi & -\sin\Delta\psi \\[3pt]
\sin\Delta\psi & \phantom{-}\cos\Delta\psi
\end{bmatrix}.
\tag{A.13}
\end{equation}
Writing $\mathbf{H}$ in components as
\begin{equation}
\mathbf{H}
=
\begin{bmatrix}
h_{11} & h_{12} \\[3pt]
h_{21} & h_{22}
\end{bmatrix},
\tag{A.14}
\end{equation}
the trace objective in Eq.~\ref{eq:trace_max} evaluates to
\begin{equation}
\operatorname{tr}\!\left(\mathbf{R}\mathbf{H}^\top\right)
=
(h_{11} + h_{22})\cos\Delta\psi
+
(h_{21} - h_{12})\sin\Delta\psi.
\tag{A.15}
\label{eq:trace_expand}
\end{equation}
This is a sinusoidal function in $\Delta\psi$. Taking its derivative with respect to $\Delta\psi$ and setting it to zero,
\begin{equation}
-(h_{11} + h_{22})\sin\Delta\psi^*
+
(h_{21} - h_{12})\cos\Delta\psi^*
= 0,
\tag{A.16}
\end{equation}
which yields the optimal rotation angle
\begin{equation}
\boxed{
\Delta\psi^*
=
\operatorname{atan2}\!\left(
h_{21} - h_{12},\;
h_{11} + h_{22}
\right).
}
\tag{A.17}
\label{eq:opt_angle}
\end{equation}
We can verify that this is indeed a maximum by checking the second derivative,
\begin{equation}
\frac{d^2}{d(\Delta\psi)^2}
\operatorname{tr}\!\left(\mathbf{R}\mathbf{H}^\top\right)
\bigg|_{\Delta\psi = \Delta\psi^*}
=
-\sqrt{(h_{11}+h_{22})^2 + (h_{21}-h_{12})^2}
< 0,
\tag{A.18}
\end{equation}
confirming that $\Delta\psi^*$ maximizes the trace objective (assuming $\mathbf{H}$ is nonzero).

\begin{algorithm}[t]
\caption{Robust SE(2) Odometry Estimation via IRLS.}
\label{alg:irls}
\begin{algorithmic}[1]
\Require Source positions $\{\mathbf{p}_i\}_{i=1}^N$, soft target positions $\{\tilde{\mathbf{p}}_i\}_{i=1}^N$, quality priors $\{\bar{w}_i\}_{i=1}^N$, Huber threshold $c$, maximum iterations $L$, convergence tolerance $\epsilon$
\Ensure Optimal rotation $\mathbf{R}^*$ and translation $\mathbf{t}^*$

\State \textbf{Initialize} $\omega_i \leftarrow \bar{w}_i$ for all $i$ \Comment{First iteration uses unit IRLS weights}
\For{$\ell = 1, 2, \ldots, L$}
    \State Compute weighted centroids $\boldsymbol{\mu}_{\mathbf{p}},\, \boldsymbol{\mu}_{\tilde{\mathbf{p}}}$ via Eq.~\ref{eq:centroids}
    \State Center the point sets $\hat{\mathbf{p}}_i \leftarrow \mathbf{p}_i - \boldsymbol{\mu}_{\mathbf{p}}$, \; $\hat{\tilde{\mathbf{p}}}_i \leftarrow \tilde{\mathbf{p}}_i - \boldsymbol{\mu}_{\tilde{\mathbf{p}}}$
    \State Compute cross-covariance $\mathbf{H} \leftarrow \sum_i \omega_i\, \hat{\tilde{\mathbf{p}}}_i\, \hat{\mathbf{p}}_i^\top$ \Comment{Eq.~\ref{eq:cross_cov}}
    \State Compute optimal angle $\Delta\psi \leftarrow \operatorname{atan2}(h_{21} - h_{12},\; h_{11} + h_{22})$ \Comment{Eq.~\ref{eq:opt_angle}}
    \State Form rotation $\mathbf{R} \leftarrow \mathbf{R}(\Delta\psi)$
    \State Compute translation $\mathbf{t} \leftarrow \boldsymbol{\mu}_{\tilde{\mathbf{p}}} - \mathbf{R}\, \boldsymbol{\mu}_{\mathbf{p}}$ \Comment{Eq.~\ref{eq:opt_t}}
    \State Compute residuals $r_i \leftarrow \left\| \mathbf{R}\mathbf{p}_i + \mathbf{t} - \tilde{\mathbf{p}}_i \right\|_2$ for all $i$
    \State Update effective weights $\omega_i \leftarrow \bar{w}_i \cdot \psi(r_i)$ via Eq.~\ref{eq:irls_weight} and Eq.~\ref{eq:effective_weight}
    \If{$|\Delta\psi^{(\ell)} - \Delta\psi^{(\ell-1)}| < \epsilon$ \textbf{and} $\|\mathbf{t}^{(\ell)} - \mathbf{t}^{(\ell-1)}\| < \epsilon$}
        \State \textbf{break} \Comment{Convergence reached}
    \EndIf
\EndFor
\State \Return $\mathbf{R}^* \leftarrow \mathbf{R}$, \; $\mathbf{t}^* \leftarrow \mathbf{t}$
\end{algorithmic}
\end{algorithm}



\subsection*{A.3\quad Complete IRLS Procedure}

Combining the results above, we summarize the complete solver in Algorithm~\ref{alg:irls}.

\end{document}